\documentclass{article}
\usepackage{ijcai25}

\usepackage{times}
\usepackage{soul}
\usepackage{url}
\usepackage[hidelinks]{hyperref}
\usepackage[utf8]{inputenc}
\usepackage[small]{caption}
\usepackage{graphicx}
\usepackage{amsmath}
\usepackage{amsthm}
\usepackage{booktabs}
\usepackage{algorithm}
\usepackage{algorithmic}
\usepackage[switch]{lineno}

\usepackage{amsfonts}

\DeclareMathOperator*{\argmax}{arg\,max}
\DeclareMathOperator*{\argmin}{arg\,min}

\newtheorem{example}{Example}
\newtheorem{theorem}{Theorem}
\newtheorem{remark}{Remark}

\title{Ordered Semantically Diverse Sampling for Textual Data}

\author{
Ashish Tiwari$^1$
\and
Mukul Singh$^1$\and
Ananya Singha$^{1}$\And
Arjun Radhakrishna$^1$\\
\affiliations
$^1$Microsoft Corp.\\
\emails
\{astiwar, singhmukul, ananyasingha, arradha\}@microsoft.com
}

\def\isnew{{\texttt{IsNew}}}
\def\wasted{{\texttt{Wasted}}}
\def\aggwasted{{\texttt{AggWasted}}}
\def\real{{\mathbb{R}}}
\def\zz{{\mathbb{Z}}}
\newcommand\str[1]{``{\texttt{#1}}"}
\newcommand\qq[1]{``{{#1}}"}
\newcommand\seq[1]{\langle {{#1}}\rangle}
\newcommand\ignore[1]{{}}

\begin{document}

\maketitle

\begin{abstract}
The goal of diversity sampling is to select a representative subset of data in a way that maximizes information contained in the subset while keeping its cardinality small.  We introduce the ordered diverse sampling problem based on a new metric that measures the diversity in an ordered list of samples.  We present a novel approach for generating ordered diverse samples for textual data that uses principal components on the embedding vectors.  The proposed approach is simple and compared with existing approaches using the new metric.  We transform standard text classification benchmarks into benchmarks for ordered diverse sampling.  Our empirical evaluation shows that prevailing approaches perform $6$\% to $61$\% worse than our method while also being more time inefficient. Ablation studies show how the parts of the new approach contribute to the overall metrics.
\end{abstract}

\section{Introduction}

Diversity sampling refers to the process of picking a representative
subset of data points from a given data set such that the samples picked
are different from each other. The goal of diversity sampling is to
capture the diversity of the data points in the dataset.
Diversity is an important need in many scenarios, but most, if not all,
work on diversity has been done in specific contexts tied to use cases.
In this paper, we take a look at diversity sampling in its own right.

Diversity sampling can be used to sample from a
learned distribution.
When picking $k$ samples to show the user from a
probability distribution generated by some algorithm or a generative model, it is
fruitful to pick a diverse set of $k$ elements rather than a set that
contains a lot of similar data points. Doing so increases
the probability of the desired data point being in the selected
sample~\cite{diverse-mrfs,diverse-beam-search,diverse-decoding}.

Diversity sampling can also be used when collecting or annotating data for training models. 
In this case, it is beneficial to ensure that the data are diverse and cover all different possibilities~\cite{wei-etal-2013-using,sener2018active}.
There are two variants here:
(1) selecting data at the beginning when there is no
(partially) trained model, and
(2) selecting data after having trained a model in previous
iterations.
The second variant is covered by active learning~\cite{settles.tr09,munro}. 
This paper focuses on the first variant.

The commonly used formalization of diversity is in terms of ``distance'' between points.
In this case, selecting a diverse sample corresponds to selecting points such that
the distance between the selected points is maximized while the
distance of the unselected points to the selected set is minimized. Such notions can 
be formalized as
a facility-location problem. These problems are easily seen to be hard, which motivates
looking at approximate solutions~\cite{diversity-nips2016,max-sum-trans}.
There are also works that sample diverse sets under additional constraints, such 
as fairness~\cite{diverse-under-fairness}, or sampling from the solution space of 
some given constraint~\cite{diverse-constraint-programming,EITER_2011}.

Diversity was used as a ``regularization term'' in the loss (reward) function
used for extractive document summarization~\cite{lin-bilmes-2011-class}.
Doing so results in the selection of sentences from {\em{different}} semantic clusters 
for inclusion in a document summary, and this diversity-based
reward led to improved summaries~\cite{lin-bilmes-2011-class}.
Generalizing beyond extractive document summarization, picking a few salient
data points from a dataset also occurs in coresets. 
A coreset is a subset of the data such that a model trained on the subset will
perform almost as well as the model trained on the full dataset~\cite{Agarwal2005coresets,feldman2020introductioncoresetsupdatedsurvey}. Coreset construction methods have been proposed for particular problems.
We want to remain problem-agnostic. Coresets have also been calculated for
efficient computation of gradients in a problem-agnostic way~\cite{10.5555/3524938.3525583}.
Just as the formalization above of diversity as facility location problems,
coreset computation problems also are hard optimization problems,
and usually greedy methods are shown to provide effective approximations, whose theory is
based on formulating submodular optimization functions~\cite{lin-bilmes-2011-class}.

Our work here is different from existing work in two important ways.
First, the aforementioned works on diverse sampling rely on a notion of distance, which typically is 
task dependent. Since we want to be task agnostic, we assume that we do not have access to any 
fixed notion of distance. Instead, we assume that there exists an oracle
$\isnew$ that determines if a point is ``new'' with respect to the previously selected points,
but access to this oracle is available only at evaluation time. 
Second, we are interested in generating an {\em{ordered}} list of samples, which we
motivate below.

We are motivated by new use cases of diverse sampling that arise due to
the availability of powerful pretrained large language models (LLMs).
One way to use LLMs is through few-shot prompting, where a small sample of examples is presented
to the LLM in the prompt. These examples are retrieved from a bank of examples, which should ideally contain a diverse set of examples that cover all possible task instances.  If a human annotates data for the bank, then
getting a {\em{small}} diverse set is important to keep annotation costs in check~\cite{alcoforado2024randominformeddataselection}.

\begin{figure}[t]
    \includegraphics[trim=0 200 400 50,clip,width=0.8\columnwidth]{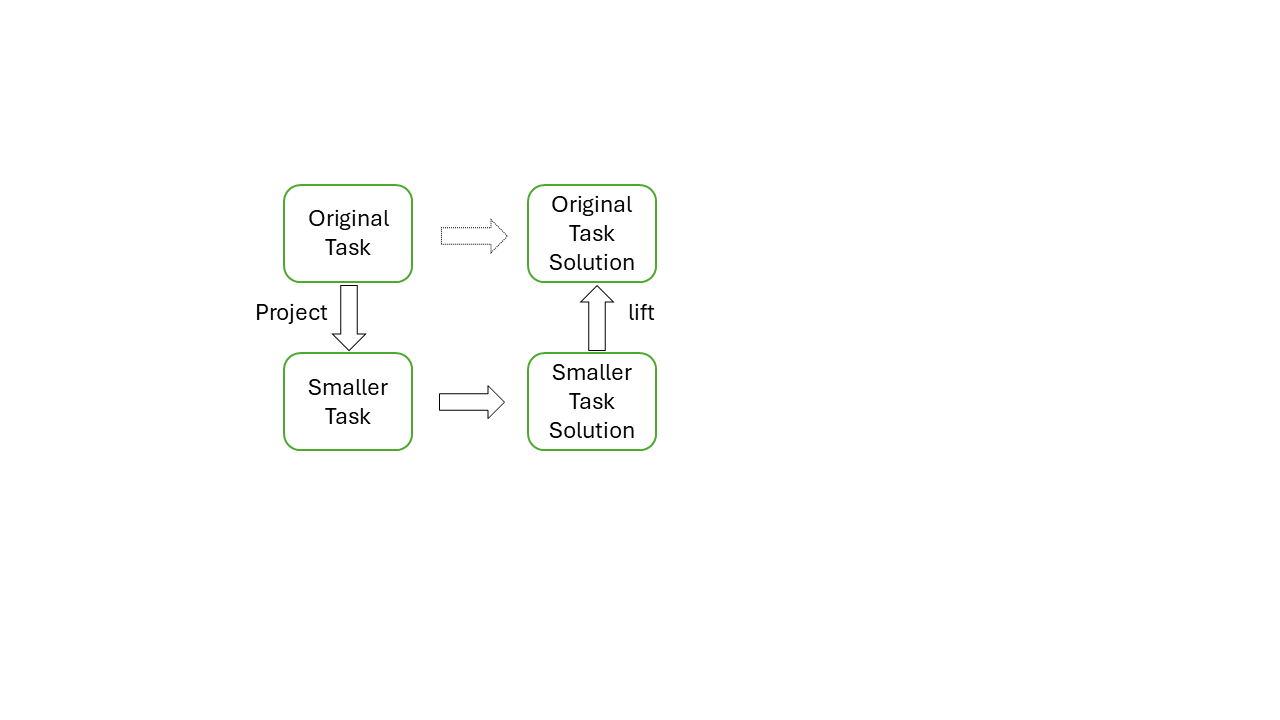}
    \caption{Motivation for Diversity Sampling}\label{fig:motivation}
\end{figure}

A second, and more important motivation, especially for ordered diverse sampling,
comes from making effective use of large-language models for large (say, data analysis) tasks. 
For many such tasks, it is not cost effective, and sometimes even prohibitive due
to prompt length constraints, to perform the {\em{full}} task using an LLM.
If we view the LLM as a good, but inefficient, way to solve a problem, we can
revisit the paradigm of coresets to effectively exploit LLMs. Recall that in the coreset paradigm,
``for efficiently approximating various extent measures of a point set $P$, 
  one quickly computes a small subset $Q$ of $P$, called a coreset, that approximates the original set $P$,
  solves the problem on $Q$ using a relatively inefficient algorithm,  and then
  translates the solution for Q to an approximate solution to the original 
  point set $P$.''~\cite{Agarwal2005coresets,10.1145/1008731.1008736}.
Using LLM as the ``relatively inefficient algorithm'', we get the following appealing paradigm for solving a data analysis
task on a large dataset: 
\\
(1) first replace the source dataset by a small dataset containing only a diverse sample, 
\\
(2) use the LLM to solve the task on the smaller sampled dataset, and finally, 
\\
(3) lift the solution to the original large dataset. 
\\
This three-step process is depicted in Figure~\ref{fig:motivation}.
As a concrete example, consider the problem of summarizing a large
dataset to the user, or labeling a large dataset consisting of 
textual data.  Using the approach in Figure~\ref{fig:motivation},
we would solve such a task by (1) getting a diverse sample,
(2) asking the LLM to solve the summarization or labeling task on the sample, and
finally, (3) lifting the solution on the sample to a solution on the original dataset. 
A diversity sample here helps by ensuring that the third step remains easy and feasible.
We use the above paradigm as our primary motivation for this work.

In this paper, we study diversity sampling in a model- and task-agnostic setting.
One immediate challenge is that there is no standard notion of diversity and hence,
no standard way for evaluating diversity samplers. 
We define a notion of diversity samples as an {\em{ordered}} sequence of data points.
We also define a metric called {\em{aggregated wasted opportunity}} that counts the number of times a diverse sample could have been picked, but was not picked (in an ordered sequence of data points).  
This metric allows us to compare samplers across benchmarks and techniques. 
We also propose a concrete approach for computing diverse samples.
We use text classification benchmarks as diversity sampling benchmarks to 
compare the various diversity (model-agnostic) sampling approaches. 
We experimentally show that our diversity samplers outperform state-of-the-art approaches.

\section{Related Work}
\label{sec:related}

Diversity sampling arises naturally in the context of active learning~\cite{settles.tr09}.
Several techniques in active learning use the model to pick samples, such as,
uncertainty sampling~\cite{10.1007/978-1-4471-2099-5_1} and
query by committee~\cite{10.1145/130385.130417}.
We focus exclusively on model-agnostic approaches.

Coresets can be used as a possible formalism for diversity sampling~\cite{Agarwal2005coresets}.
Active learning itself can also be framed as a coreset selection problem~\cite{sener2018active}.
At the core, the problems can be mapped to finding a subset $S$ from a (data)set $D$ that maximizes
the (coverage) $F(S)$ -- computed using a function $F: 2^D \mapsto \real$ -- and subject to $|S| \leq k$.
Such problems are typically NP-hard, but admit efficient approximation algorithms when
$F$ satisfies some conditions~\cite{lin-bilmes-2011-class}. These efficient algorithms take the
form of greedy algorithms. Efficiency is very important since sizes of the dataset $D$ can be large.

Specifically, for diversity sampling, we have the following model-agnostic approaches
described in the literature. The first class of approaches are {\em{clustering-based}}, where the
dataset $D$ is partitioned into clusters (using an unsupervised clustering technique) and
samples are picked from each cluster. The samples are picked so that they are either close to the center of
a cluster, close to the edge, or randomly placed~\cite{munro}.
A second baseline is the {\em{greedy algorithm for the k-center problem}}, which works by starting
with a randomly selected data point and putting it in $S$, and thereafter iteratively selecting
points farthest away from the current set $S$, and adding it to $S$~\cite{sener2018active}.
A third baseline is the {\em{reverse semantic search}} procedure that works by starting with
$S = \emptyset$, and thereafter iteratively selecting two data points that are farthest 
away from each other among the points in $D\setminus S$ and adding them to $S$~\cite{alcoforado2024randominformeddataselection}.
We compare our proposed approach with all these three baselines mentioned above.

We note that the paper~\cite{alcoforado2024randominformeddataselection} notes that
it is very difficult to beat random sampling especially on datasets that do not have 
outliers.  They use the metric ``overannotation rate'', which is similar in spirit to our metric, 
but it can not be reasonably aggregated across benchmarks. 
The paper also proposes a few other approaches, but reverse random sampling is reported
as the overall best approach, along side random sampling. We also compare against random
sampling.

\ignore{
\cite{sener2018active}:
This problem considers a fully labeled dataset and tries to choose a subset of it such that the model trained on the selected subset will perform as closely as possible to the model trained on the entire dataset. For specific learning algorithms, there are methods like core-sets for SVM (Tsang et al., 2005) and core-sets for k-Means and k-Medians (Har-Peled and Kushal, 2005). However, we are not aware of such a method for CNNs.
The most similar algorithm to ours is the unsupervised subset selection algorithm in (Wei et al., 2013). It uses a facility location problem to find a diverse cover for the dataset. Our algorithm differs in that it uses a slightly different formulation of facility location problem. Instead of the min-sum, we use the minimax (Wolf, 2011) form. More importantly, we apply this algorithm for the first time to the problem of active learning and provide theoretical guarantees for CNNs.

k-center greedy: start with a data point, pick the farthest from the selected set until you hit budget.
\endignore}

%

\section{Diversity Sampling Problem and Metric}

In this section, we will fix the notation, introduce a new metric to 
define the diversity sampling problem.

Let $D$ be a dataset consisting of textual data. Formally, 
$D$ is an $(N \times 1)$-matrix of string values.
Our goal is to get a small sample of $n$ values from $D$ where $n \ll N$ that is \qq{\em{diverse}}. We first need to define a metric that will be a measure of diversity.

Diversity is a relative concept that is dependent on a notion of what is 
\qq{same} and what is \qq{different} in a sequence of values.
Let us assume we have a function $\isnew$ that captures this notion and takes an ordered sequence of data points from $D$, say 
$\seq{d_1,d_2,\ldots,d_k}$, and returns either 1 or 0.
Intuitively, $\isnew$ returns 1 if $d_k$ is a new (diverse) data point that is 
different from all previous ones $d_1,\ldots,d_{k-1}$ occurring before it in 
the ordered sequence.

In our formalism, we will treat diverse samples as {\em{ordered sequences}}. 
Intuitively, we want to cover as much diversity as possible in the initial segments of the ordered sequence.
The motivation is that if there is a limited budget, say $k$, for the number of samples, an ordered sequence allows us
to pick the first $k$ elements as the diverse sample set.
If $\seq{d_1,d_2,\ldots,d_k}$ is a diverse sample, we expect
$\isnew(\seq{d_1,\ldots,d_i})$ to return 1 for every $i$ 
in $\{1,\ldots, k\}$. This suggests that we can define a metric that just adds up
the values returned by $\isnew$ on every prefix subsequence of a given
sequence to define the measure of diversity of that sequence.
However, there are two issues with this measure.
(1) This measure penalizes a sequence even
when there {\em{no more new samples}} to pick for its last element, and
(2) This measure does not reward sequences that get all the diverse
elements early.
Let us illustrate the above issues with an example.

\begin{example}\label{ex:one}
    Consider a toy dataset 
    $D = [ \str{1A}, \str{1B}, \str{2B}, \str{2C}, \str{1A}, \str{2B} ]$ thought of as a $(6\times 1)$ matrix of strings. 
    Let $\isnew(\seq{d_1,d_2,\ldots,d_k})$ 
    be a function that returns $0$ if the first character of the string $d_k$ was also the first character of some string $d_i$ for $i < k$. 
    Thus, $\isnew(\seq{\str{1A}, \str{1B}}) = 0$, whereas 
    $\isnew(\seq{\str{1A}, \str{1B}, \str{2B}}) = 1$.
    Let us say a diversity sampler
    picked the indices $\seq{0,1,2}$, which correspond to
    the strings $D_1 := \seq{\str{1A}, \str{1B}, \str{2B}}$, 
    and a second diversity
    sampler picked up the indices $\seq{0, 2, 3}$, which correspond to 
    the strings $D_2 := \seq{\str{1A},\str{2B},\str{2C}}$.
    If we sum up the $\isnew$ scores for each prefix subsequence of $D_1$, we get $1 + 0 + 1 = 2$, and if we do the same for $D_2$, we get $1 + 1 + 0 = 2$.
    Note that $D_2$ picked the perfect diversity sample (with respect to
    $\isnew$) in the first two selections, but got penalized with a $0$ for the third selection even though there was no better choice to make there.
    On the other hand, $D_1$ picked the \qq{wrong} string as its second selection, which rightly causes it to get a score of $0$ for that choice, but it still gets the same overall total score. \qed
\end{example}

We fix these two issues by defining a metric that counts the number of 
\qq{wasted opportunity}; that is, the number of times there was an opportunity to pick a diverse sample, but instead a non-diverse candidate was picked. Formally,
given the sequence
$\seq{d_1, d_2, \ldots, d_k}$,
the pick $d_k$ is a {\em{wasted opportunity}}, denoted
$\wasted(\seq{d_1,\ldots,d_k}) = 1$, if \\
(a) $\isnew(\seq{d_1,d_2,\ldots,d_k}) = 0$, and \\
(b) there exists $d_j\in D$ s.t.
    $\isnew(\seq{d_1,d_2,\ldots,d_j}) = 1$. \\

Given a sequence $\seq{d_1,d_2,\ldots,d_k}$, the {\em{aggregated wasted opportunity}} metric is a number from $0$ to $k$ that is defined by
\begin{eqnarray*}
    \aggwasted & := & \sum_{i=1}^{k} \wasted(\seq{d_1,\ldots,d_i})
\end{eqnarray*}
In other words, aggregated wasted opportunity is just a count of the picks $d_i$ that are a wasted opportunity with respect to the prior picks $\seq{d_1,\ldots,d_{i-1}}$. A smaller aggregated wasted opportunity score indicates a higher quality diversity sample.

\begin{example}\label{ex:two}
    Continuing the running example from Example~\ref{ex:one}, for the first sequence,
    we note that
    $\wasted(\seq{\str{1A}, \str{1B}}) = 1$, but
    $\wasted(\seq{\str{1A}, \str{1B}, \str{2B}}) = 0$,
    and so the aggregated score for the full sequence is
    $\aggwasted(\seq{\str{1A}, \str{1B}, \str{2B}}) = 0 + 1 + 0 = 1$.
For the second sequence, we note that
    $\wasted(\seq{\str{1A}, \str{2B}}) = 0$, and
    $\wasted(\seq{\str{1A}, \str{2B}, \str{2C}}) = 0$ because
    there is no other choice $d_j$  s.t.
    $\isnew(\seq{\str{1A},\str{2B},d_j}) = 1$.
    Thus, for the second sequence, the aggregated score
    $\aggwasted(\seq{\str{1A}, \str{2B}, \str{2C}})$ is
    $0 + 0 + 0 = 0$. The new aggregated wasted opportunity
    metric is able to distinguish between the two diverse sample
    sequences. \qed
\end{example}

Now, we are ready to define the diversity sampling problem.
Given a dataset $D$ consisting of data points $d_1,\ldots,d_n$,
the goal of diversity sampling is to select an {\em{ordered}} list of
indices $i_1,i_2,\ldots,i_k$ such that the aggregated wasted
opportunity score, given by
$\aggwasted(\seq{d_{i_1},d_{i_2},\ldots,d_{i_k}})$, is minimum.
The underlying function $\isnew$ is not given a-priori, and should
be assumed to be some \qq{natural} notion of categorizing data points
as different or similar to those seen before.

\begin{example}\label{ex:three}
    Going back to Example~\ref{ex:one},
    we had used one notion of $\isnew(\seq{d_1,\ldots,d_k})$ 
    that decided what two data points were similar based on the first character of the strings. Another reasonable candidate could be a function
    $\isnew$ that decides based on the second character, which will
    make $\str{1B}$ similar to $\str{2B}$ and different from $\str{1A}$.
    Now, if we use this new function $\isnew$ to define our aggregated
    wasted opportunity metric, and use it compare the same two sample sequences
    from Example~\ref{ex:one}, we get that
    $\aggwasted(\seq{\str{1A}, \str{1B}, \str{2B}}) = 0 + 0 + 1 = 1$, and
    $\aggwasted(\seq{\str{1A}, \str{2B}, \str{2C}}) = 0 + 0 + 0 = 0$.
    So, even under this different notion of similarity, the diversity
    sampler that returns the second sequence is deemed better in our 
    metric. Note that in the first sequence, the choice $\str{2B}$ 
    is a wasted opportunity since there exist other diverse
    candidate strings,
    such as $\str{2C}$, that could have been picked. \qed
\end{example}

\subsection{Benchmarks for Evaluating Diversity Samplers}
Having fixed the metric that can distinguish between good and bad diversity samplers,
the next natural question is what benchmarks do we use to evaluate diversity samplers.
A good source for benchmarks 
are the {\em{classification benchmarks}}.  In a classification benchmark,
a dataset $D$ is associated
with a labeling function $L: D \mapsto C$, where $C$ is a finite set of
categories. We can use the labeling function $L$ to  naturally define the $\isnew$ function as follows:
\begin{eqnarray}
    \isnew(\seq{d_1,\ldots,d_k}) & := & 
    \left\{ \begin{array}{l} 0 \quad \mbox{if $\exists{i<k}: L(d_k) = L(d_i)$}
        \nonumber
    \\ 1 \quad \mbox{otherwise} \end{array} \right. \label{eqn:isnew}
\end{eqnarray}
Intuitively, a sample is new with respect to previously picked samples if its associated label is new and not seen before.
Thus, each classification benchmark naturally transforms to a diversity sampling benchmark. In our experimental evaluation, we compare diversity sampling algorithms using the aggregated wasted opportunity metric (with respect to the 
$\isnew$ function induced by the labeling) on such benchmarks.

\begin{remark}
The wasted opportunity metric has the nice property that it remains comparable across benchmarks that have different number of labels.
Consequently, we can aggregate this metric over different benchmarks, which in turn allows us to compare efficacy of different diversity 
samplers over a wide array of benchmarks. \qed
\end{remark}

\begin{remark}
In our formalism, the $\isnew$ function is not constrained in any way, and in fact, we can make it quite interesting.
For example, consider the set $\{1,2,\ldots,7\}$ of 7 data points, and the 
    $\isnew$ function defined as:
    $\isnew(\seq{d_1,\ldots,d_k})$ is $0$ iff $\exists{i<k}: |d_i-d_k| < 3$, and it is $1$ otherwise.
    For this function, the sequence $\seq{1,4,7}$ has only \qq{new} elements since all elements are at least $3$ units apart, and hence its wasted opportunity metric is $0$.
    However, if we pick $\seq{1,5}$ as the first two elements, then there is no choice for the third element that is \qq{new}. If we insist that diversity samples sequences be of length $3$, then picking $1,5$ will cause us to necessarily end up with a 
    wasted opportunity metric of $1$. Picking $1,4$ would have retained the option of picking $7$ and still keeping the
    wasted opportunity metric to $0$. 
    While such $\isnew$ functions are possible, the one we target in this paper and also use in our evaluation, which is
    defined in Equation~\ref{eqn:isnew}, does not cause such complications.
    \qed
\end{remark}

\section{Principled Samplers}

We now describe our high-level approach for selecting a diverse sample
from a given textual dataset $D$.

The intuition behind our procedure is rather simple. If we embed the
text into an $m$-dimensional real space, $\real^m$, and then look at the
principal components of the data in $\real^m$, the data points that have
a large (positive and negative) projection 
along the principal components will be \qq{different} from each other and good
candidates for inclusion in a diverse sample.  Also, the data points that
have low projections on all the principal components will also be 
qualitatively different from those picked above, and good candidates for
inclusion. 

The algorithm {\texttt{Principled Sampler v1}} 
is presented in Algorithm~\ref{alg:v1}.
The input to the algorithm is a dataset $D$ consisting of text column
with, say, $N$ rows.
The parameters to the algorithm are an embedding function $E$ and an
integer $n$. The sampler works by first computing the embedding for each
of the $N$ strings to get a real matrix $X$ with shape $N \times M$, where
$M$ is the dimension of the embedding space.
We then use an off-the-shelf package for 
principal component analysis (PCA), and find the $n$ principal
components and project the data in $X$ to those components to get
$X_{pca}$ which is a real matrix with shape $N \times n$.
Let us use the notation
$X_{pca}[j,i]$ for the element in $j$-th row and $i$-th column in the matrix $X_{pca}$.
Let 
$X_{pca}[*,i]$ denote the $i$-th column of $X_{pca}$ ($i\in\{1,\ldots,n$)
thought of as a vector of size $N$.
Similarly, let
$X_{pca}[j,*]$ denote the $j$-th row of $X_{pca}$ ($j \in \{1,\ldots,N\}$)
thought of as a vector of size $n$.
We first pick the index $j$ of the vector $X_{pca}[*,1]$ that has the largest value 
$X_{pca}[j,1]$ among all the values in this vector. 
We do the same for all the columns of $X_{pca}$ and get the index of the largest value
in each column. These indices are stored in $Y$.
We next pick the index $k$ of the vector $X_{pca}[*,1]$ that has the smallest value 
$X_{pca}[k,1]$ among all the values in this vector. 
We do the same for all the columns of $X_{pca}$ and get the index of the smallest value
in each column. These indices are stored in $Z$.
Finally, we compute the infinity norm of the rows of $X_{pca}$ and store them 
in $X_\infty$. We then pick the $n$ indices that contain the $n$ smallest values
in $X_\infty$. These indices are stored in $W$.
Recall that the infinity norm of a vector is just the maximum of the absolute values of
all components of the vector.
Algorithm~\ref{alg:v1} returns the concatenation of the arrays $Y, Z$ and $W$. The
return value is an array containing $3n$ indices of data points that are included
in the diversity sample.

The $n$ indices in array $Y$ correspond to data points that have the largest projection on
the $n$ principal components respectively.
The $n$ indices in array $Z$ correspond to data points that have the smallest (most negative) projection on 
the $n$ principal components respectively.
The $n$ indices in array $W$ correspond to data points that have the $n$ smallest absolute projection on each of 
the $n$ principal components.
Our procedure returns the $3n$ samples containing the union of the three arrays.
The intuition is that the first two sets $Y$ and $Z$ contain indices of data points that capture the
\qq{common} pattern in the dataset $D$,
whereas the last set $W$ contains the indices of \qq{uncommon} pattern 
and \qq{outliers}~\cite{fariha2021conformance}.
Note that in the returned ordered sequence, $Y$ comes before $Z$ and $Z$ comes before $W$.

\begin{algorithm}[tb]
    \caption{Principled Sampler v1}
    \label{alg:v1}
    \begin{algorithmic}[1] 
        \REQUIRE {\em{Input}}: data $D$, an $(N\times 1)$-matrix of strings
        \REQUIRE {\em{Parameters}}: embedder $E$, integer $n$
        \ENSURE  {\em{Output}}: Ordered sequence of $3n$ indices
        \STATE $X \gets E(D)$ \COMMENT{$X \in \real^{N\times M}$}
        \STATE $X_{pca} \gets PCA(X, \mbox{components}=n)$ \COMMENT{$X_{pca}\in\real^{N\times n}$}
        \STATE Initialize arrays $Y,Z$ of size $n$ each  \COMMENT{$Y,Z\in \zz^{1\times n}$}
        \FOR{$i \in \left[1,2,\ldots,n\right]$}
         \STATE $Y[i] \gets j$ where $X_{pca}[j,i] = \max( X_{pca}[*,i] )$
         \STATE $Z[i] \gets k$ where $X_{pca}[k,i] = \min( X_{pca}[*,i] )$
        \ENDFOR
        \FOR{$j \in \left[1,2,\ldots,N\right]$}
         \STATE $X_\infty[j] \gets ||X_{pca}[j,*]||_\infty$ \COMMENT{$X_\infty\in\real^{1\times N}$}
        \ENDFOR
        \STATE $W$ are indices of the $n$ smallest values in $X_\infty$
        \RETURN $[ Y, Z, W ]$ \COMMENT{Concatenate the 3 arrays}
    \end{algorithmic}
\end{algorithm}

\begin{remark}
    It is possible that the indices in $Y$, $Z$ and $W$ overlap, which can cause
    Algorithm~\ref{alg:v1} to return less-than $3n$ samples. For evaluation purposes, we want all
    techniques to return the same number of samples. To force Algorithm~\ref{alg:v1} to return
    exactly $3n$ samples, we modify it as follows: whenever a repeated value is generated in either 
    $Y$, $Z$ or $W$,
    we replace that value by the next best pick for that set. \qed
\end{remark}

\subsection{Principled Sampler v2}

We now describe a slightly modified version of the 
\qq{Principled Sampler v1} described above, which we call
\qq{Principled Sampler v2}.

The algorithm {\texttt{Principled Sampler v2}} 
is presented in Algorithm~\ref{alg:v2}.
The changes are quite minimal. In Principled Sampler v1,
we picked the data points that had extreme projections on the
principal components.  In Principled Sampler v2, we pick
data points that each have a very large projection on one of the
principal components, {\em{and a very small projection on
the other components}}. 
In the notation used in Algorithm~\ref{alg:v2},
the expression 
$X_{pca}[j,i] - \sum_{k\neq i}|X_{pca}[j,k]|$ 
is maximized when $X_{pca}[j,i]$ is large and $X_{pca}[j,k]$ is close to zero for all $k \neq i$.
Algorithm~\ref{alg:v2} picks $j^*$ that maximizes this expression and these $j^*$ (one for each choice of $i$)
are included in $Y$.
Similarly, 
the expression
$X_{pca}[j,i] + \sum_{k\neq i}|X_{pca}[j,k]|$ 
is minimized when $x_j[i]$ is 
small and $X_{pca}[j,k]$ is close to zero for all $k \neq i$.
Algorithm~\ref{alg:v2} picks $j^*$ that minimizes this expression, and these $j^*$ (one for each choice of $i$)
are included in $Z$.
The outlier picks in $W$ are identical to those in Algorithm~\ref{alg:v1}.

\begin{algorithm}[tb]
    \caption{Principled Sampler v2}
    \label{alg:v2}
    \begin{algorithmic}[1] 
        \REQUIRE Input, parameters same as in Algorithm~\ref{alg:v1}
        \ENSURE Output same as in Algorithm~\ref{alg:v1}
        \STATE $X \gets E(D)$
        \STATE $X_{pca} \gets PCA(X, \mbox{components}=n)$
        \FOR{$i \in \left[1,2,\ldots,n\right]$}
         \STATE $Y[i] \gets j^*$ where $j^* = \argmax_j( X_{pca}[j,i] - \sum_{k\neq i}|X_{pca}[j,k]| )$
         \STATE $Z[i] \gets j^*$ where $j^* = \argmin_j( X_{pca}[j,i] + \sum_{k\neq i}|X_{pca}[j,k]| )$
        \ENDFOR
        \STATE $W$ same as in Algorithm~\ref{alg:v1}
        \RETURN $[ Y, Z, W ]$ \COMMENT{Concatenate the 3 arrays}
    \end{algorithmic}
\end{algorithm}

This variant makes more effort to ensure that the picked candidates are different from each other.
The two versions above can be viewed as instantiations of a more general 
approach, where we pick the candidates based on some ranking functions
defined over the projections on the dataset on its principal components.
We leave the exploration of other choices for ranking functions for future work and
evaluate these two concrete versions in this paper.

\begin{remark}
    Our proposed diversity sampling procedures are {\em{non-probabilistic}} and {\em{task-agnostic}}.
    By non-probabilistic, we mean that once the input dataset is fixed and the parameters are fixed, 
    the procedures return
    a fixed sequence of samples every time. 
    The two procedures are also task-agnostic, which means that 
    they return the same samples irrespective of the underlying $\isnew$ concept of
    interest.  
    It is possible to make the procedures probabilistic by replacing deterministic choices (of the
    maximum or minimum) by probabilistic choices (biased towards picking maximum, respectively minimum,
    more often). 
    It is also possible to make the procedures task-aware; for example, by incorporating information about the
    desired notion of diversity into the choice of principal components. These extensions are promising
    directions for future work. \qed
\end{remark}

\ignore{
Here a ranking function is just a function that maps each row in 
$X_{pca}$ to a real number.
If $f_1, f_2, f_3$ are three ranking functions, then the 
Generic Principled Sampler can be described as in Algorithm~\ref{alg:generic}.

\begin{algorithm}[tb]
    \caption{Principled Sampler Generic}
    \label{alg:generic}
    \textbf{Input}: data\\
    \textbf{Parameter}: embedder, n, $f_1,f_2,f_3$\\
    \textbf{Output}: Ordered 3*n samples from data
    \begin{algorithmic}[1] 
        \STATE Use embedder to compute embeddings X of data
        \STATE Use PCA to project X into n-dimensional space Xpca
        \FOR{$i \in \left[1,2,\ldots,n\right]$}
         \STATE \texttt{yield} $\argmax_{x\in Xpca} f_1(x_i,(x_j)_{j\neq i})$
         \STATE \texttt{yield} $\argmin_{x\in Xpca} f_2(x_i,(x_j)_{j\neq i})$
         \STATE \texttt{yield} index that gives $i$-th smallest value for $f_3(x_1,x_2,\ldots,x_n)$
        \ENDFOR
    \end{algorithmic}
\end{algorithm}
\endignore}

\section{Evaluation}

We evaluate the proposed 
diversity sampling procedures. As mentioned earlier, we turn
text classification benchmarks into diversity sampling benchmarks.
We use aggregated wasted opportunity as the metric for
comparing diversity samplers.

\subsection{Datasets}

We used text classification datasets to evaluate diversity samplers. 
Given a text classification dataset, the goal of a diversity sampler 
is to sample as few data points as possible from the dataset that together 
map to all possible labels. 
We used text classification datasets taken from
Papers with code~\cite{paperswithcode}, 
Kaggle~\cite{kaggle}, and Hugging Face~\cite{huggingface}.
We picked benchmarks that had a large number of labels. In total, we selected 22 benchmarks with the number of labels ranging from $3$ to $98$ and standard deviation $19.6$. We will make the list of selected benchmarks available.
\ignore{specific sources for a large subset of the datasets used are listed below:
{\small{
\url{hf://datasets/coastalcph/lex_glue/},
\url{hf://datasets/google/frames-benchmark/},
\url{hf://datasets/dair-ai/emotion/},
\url{hf://datasets/google-research-datasets/go_emotions/},
\url{hf://datasets/nyu-mll/multi_nli/} (2),
\url{hf://datasets/nvidia/Aegis-AI-Content-Safety-Dataset-1.0/},
\url{hf://datasets/declare-lab/HarmfulQA/data_for_hub.json} (2),
\url{hf://datasets/fancyzhx/ag_news/},
\url{hf://datasets/cardiffnlp/tweet_eval/},
\url{hf://datasets/zeroshot/twitter-financial-news-topic/},
\url{hf://datasets/SetFit/bbc-news/},
\url{hf://datasets/ttxy/emotion/},
\url{hf://datasets/zeroshot/twitter-financial-news-sentiment/},
\url{hf://datasets/jonathanli/law-stack-exchange/},
\url{hf://datasets/allenai/prosocial-dialog/},
\url{https://www.kaggle.com/datasets/urbanbricks/wikipedia-promotional-articles}, 
\url{https://www.kaggle.com/datasets/yelp-dataset/yelp-dataset/data},
\url{https://www.kaggle.com/datasets/hijest/genre-classification-dataset-imdb},  
\url{https://www.kaggle.com/datasets/arplusman/papers-by-subject},
\url{https://www.kaggle.com/datasets/jp797498e/twitter-entity-sentiment}
}}
\endignore}

For each of the 22 datasets, we pick the validation or test set (because
they are small, but we could have picked the train set as well). 
For each individual experiment, we further down sample, and pick a random sample of 
size $N=250$ and the ground-truth labels for those $N$ data points, to get {\em{one}} benchmark for diversity samplers.
We repeat our experiments multiple times with different seeds for the random number generator.
Recall that the $\isnew$ function for these experiments is defined by
Equation~\ref{eqn:isnew}, which uses the ground-truth labels to decide
if the $k$-th sample is different from the previous $k-1$ samples.

\subsection{Comparing with Baselines}

We first compare our proposed diversity samplers with the baseline
diversity samplers mentioned in Section~\ref{sec:related}. We only use model-agnostic and task-agnostic baselines since
they are directly comparable with our approach. In every baseline and even in our method, the first step remains unchanged: we use an embedding function, $E$, to embed the input $N$ strings in an $M$-dimensional space over reals.

\begin{figure}[t]
    \includegraphics[width=0.9\columnwidth]{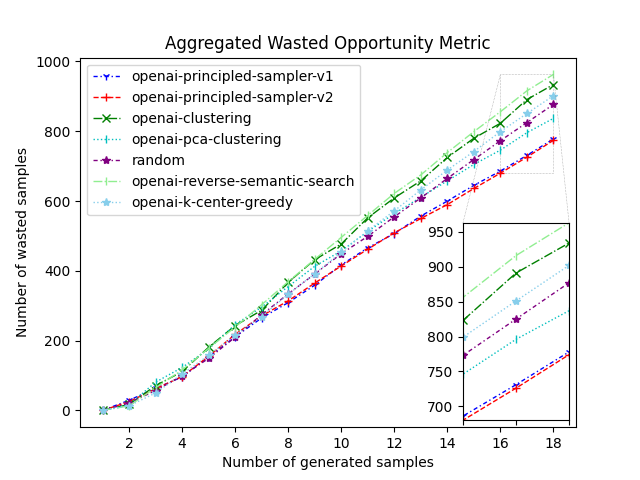}
    \caption{Comparing with  baselines using $3n=18$ samples using openai embeddings.}
    \label{fig:baseline}
\end{figure}

The first baseline, {\em{clustering}} uses the $k$-means algorithm to cluster the $N$ data points into $n$ clusters,  
and from each cluster, it picks $3$ points: (a) the point closest to the centroid of the cluster, (b) the point in the cluster farthest away from the centroid, and (c) a random point in the cluster. After processing the $n$ clusters, we get $3n$ data points in the sample set.

The second baseline, {\em{pca-clustering}}, is a variant of the above procedure where
dimension-reduction using PCA is performed before using $k$-means clustering. 

The third baseline, {\em{reverse-semantic-search}}, repeatedly picks the two points that are farthest away in the embedding space among the remaining points. After $3n/2$ iterations, we get a sample of size $3n$.

The fourth baseline, {\em{kcenter}}, starts with a random point and then iteratively picks the point that is farthest away from the current set of sampled points. After $3n-1$ iterations, we get a sample of size $3n$.

The fifth baseline, {\em{random}}, returns a random sample of size $3n$.
We use two different embeddings: {\em{openai}} refers to the use of the pretrained model, \texttt{text-embedding-3-small}, from OpenAI, and 
{\em{tfidf}} refers to the use of the TFIDF vectorizer from the sklearn library.
Specifically, we used the TFIDF vectorizer available in
sklearn library, namely {\texttt{TfidfVectorizer(max\_df=0.5, min\_df=5, stop\_words="english")}}.

Figure~\ref{fig:baseline} shows the aggregated-wasted-opportunity metric on the $y$-axis
plotted against the number $3n$ of samples on the $x$-axis. Recall that all algorithms generate
$3n=18$ samples, which are ordered. Thus, the $y$-axis is just the running total of the missed
opportunities until the $k$-th sample, as $k$ ranges from $1$ to $18$ on the $x$-axis.
Each line in the plot corresponds to a different sampler.
We have 7 lines: 5 for the 5 baselines mentioned above, and 2 for our \qq{principled sampler v1} and \qq{principled sampler v2}.
A lower value is better in the plot since a small value for aggregated-waster-opportunity indicates that the
algorithm is picking diverse samples at most of the steps where it is possible to do so.

Figure~\ref{fig:baseline} shows the approaches proposed in this paper perform better than
{\em{all}} the baselines. The \qq{pca-clustering} baseline performs closest to our
proposed method, and in fact, one can argue that performing PCA and then using unsupervised
clustering can potentially achieve the same effect as our technique. Nevertheless, our
technique is more direct and avoids clustering, and also performs consistently better.

\begin{figure}[t]
    \includegraphics[width=0.9\columnwidth]{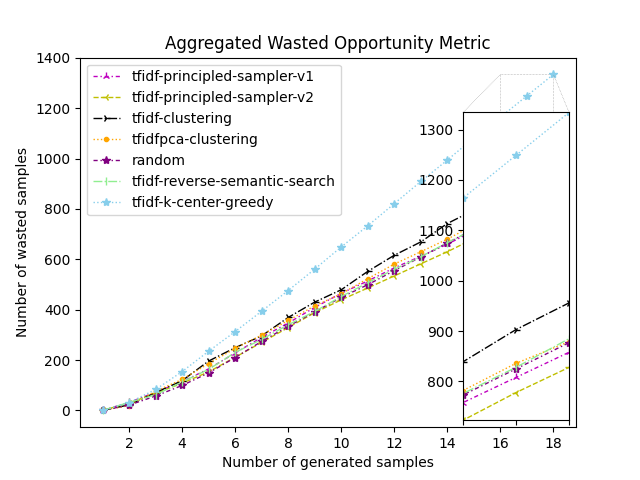}
    \caption{Comparing with  baselines using $3n=18$ samples using tfidf embeddings.}
    \label{fig:tfidf}
\end{figure}

Figure~\ref{fig:tfidf} shows the same plot as Figure~\ref{fig:baseline}, but using tfidf embeddings instead. The same observations hold true for both openai and tfidf embeddings. This shows our proposed technique is 
not just designed to work with one kind of embedding but the improvements persist across
the choice of embedding. This is further evidence that there is inherently some value in the proposed way for diversity sampling.

We note that some of the
baselines perform worse than random sampling, but our approach performs better than it. We further note that adding dimension-reduction via PCA before
clustering makes it better than random sampling. This shows that PCA helps and we believe that our approach is able to fully exploit the benefits of PCA to generate diverse samples.

\subsection{Effect of the Embedding Procedure}

In our next experiment, we evaluate the effect of the choice of the embedding.
The embedding computed by the pretrained embedding model, \texttt{text-embedding-3-small}, from OpenAI
is good, but expensive compared to the cheaper TFIDF-based embeddings that just rely on the count of words in the text. 

\begin{figure}[t]
    \includegraphics[width=0.9\columnwidth]{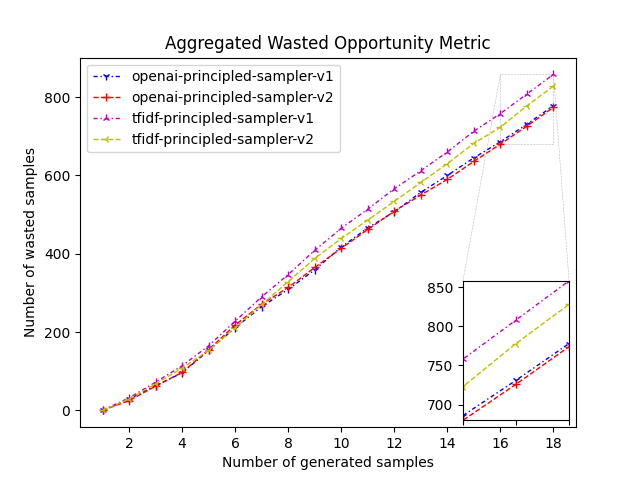}
    \caption{Comparing Samplers v1 and v2 with openai embedding and with TFIDF-based embedding.}
    \label{fig:embedding}
\end{figure}

Figure~\ref{fig:embedding} shows a plot comparing the versions of both our procedures with 
the two embedding procedures.  We see that the text-embedding model from OpenAI performs better
expectedly. We also note that our two variants, v1 and v2, perform equally well, and changing the embedding
affects both of them the same way, although version v2 appears slightly more robust to embedding change.

\begin{table}[tbh]
    \centering
    \begin{tabular}{lccccccc}
        \hline
          & v1 & v2 & cls & pca-cls & rss & k-ctr & rnd 
        \\
        \hline
        tfidf &  3.6 & 0.0 & 15.5 & 6.2 & 6.9 & 61.1 & 5.9
        \\
        openai & 0.5 & 0.0 & 20.7 & 8.1 & 24.4 & 16.5 & 13.3
        \\
        \hline
    \end{tabular}
    \caption{Percent increase in aggregated wasted opportunity scores for different approaches
    when compared to Principled Sampler v2 using OpenAI text embedding.}
    \label{tab:numbers}
\end{table}

Table~\ref{tab:numbers} shows the percentage increase in the aggregated-wasted-opportunity metric for the $7$ diversity samplers for each of the two embeddings. Since Principled Sampler $v2$ performed the best for both embeddings, the percentage increases in each row are reported with respect to $v2$'s score in that row. Existing approaches show a degradation ranging from $6\%$ to $61\%$ compared to performance of Principled Sampler $v2$. The PCA+Clustering (pca-cls) approach was within $6$-$8$\% of our approach and random (rnd) was withng $6$-$13$\%, and these two were the best among the rest. Both reverse semantic search (rss) and greedy k-center (k-ctr) are impacted a lot by the choice embedding.




\ignore{
\subsection{The Complete Plot}

TODO: Merge with baseline
\begin{figure}[t]
    \includegraphics[width=0.9\columnwidth]{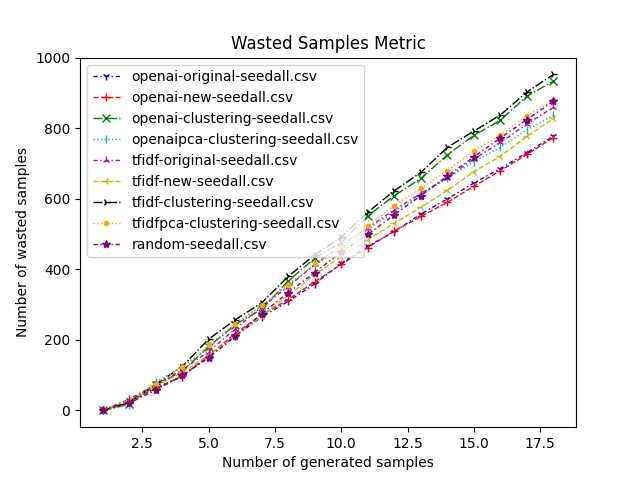}
    \includegraphics[width=0.9\columnwidth]{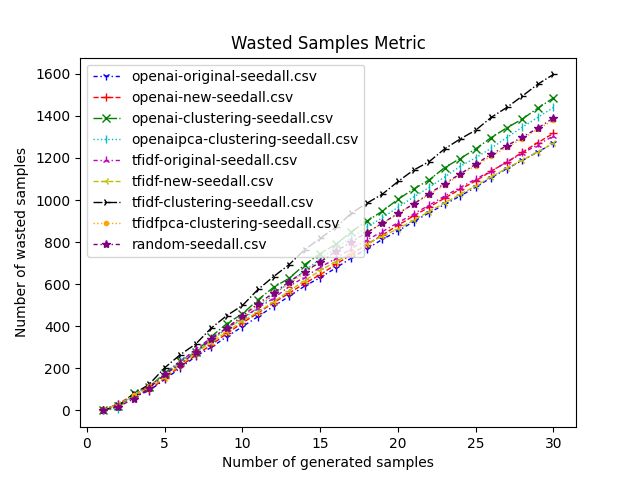}
    \caption{All results}
    \label{fig:all}
\end{figure}

Figure~\ref{fig:all} collects all variants described above, and includes a new one --
purely random sampling -- and plots them all on the same plot.  
\endignore}

\subsection{Ablation Studies}

\begin{table}[tb]
    \centering
    \begin{tabular}{lcccc}
        \toprule
        embedding & $v1$ & $v1 - Y$ & $v1 - Z$ & $v1 - W$ \\
        \midrule
        tfidf & $0.0$ & $2.8$ & $3.5$ & $-0.7$         \\
        openai & $0.0$ & $1.7$ & $7.3$ & $0.9$        \\
        \bottomrule
    \end{tabular}
    \caption{Ablation studies for Principled Samplers v1 showing the percent increase in aggregated wasted opportunity metric for each of the three ablated variants relative to corresponding baseline.}
    \label{tab:ablationv1}
\end{table}

We now evaluate the contribution of each of the three pieces of our Principled Samplers: the first sample set $Y$ that contains data points that maximize projections on a principal component, the second set $Z$ that minimizes that projection, and the third set $W$ that minimizes the absolute value of projections on all first $n$ components. We turn-off each piece and replace it by a set generated by random sampling to get versions $-Y$, $-Z$, and $-W$ respectively.
Thus, when we start with Principled Sampler $v1$ as the baseline, we get 
ablated versions $v1 - Y$, $v1 - Z$, and $v1 - W$; and
when we start with Principled Sampler $v2$ as the baseline,
we similarly get $v2 - Y$, $v2 - Z$, and $v2 - W$.

Tables~\ref{tab:ablationv1} and~\ref{tab:ablationv2} report the {\em{degradation}} in performance of each of the ablated versions compared to its {\em{respective}} baseline. Note that we perform the ablation studies for both embeddings: the one computed using the openai embedding model and the tfidf technique.
The main observation is that the aggregated wastage metric shows up to a $9.8\%$ increase in the ablated versions. There are two cases where the ablated version performed better, but only very slightly -- these are $v1 - W$ and $v2 - W$ when using tfidf as the embedding. However, the degradation is very small in both cases.  
In general, we observe that there are very small changes in performance for the $-W$ ablated versions. This is expected since the set $W$ was designed to catch outliers and non-conforming data points, but these scenarios occur infrequently in the text classification benchmarks.
It is surprising to see the imbalance between the $-Y$ and $-Z$ versions since one would expect them to perform similarly -- when a vector $v$ is picked as a principal component, $-v$ could have as well been picked.

\begin{table}[tbh]
    \begin{tabular}{lcccc}
        \toprule
        embedding & $v2$ & $v2 - Y$ & $v2 - Z$ & $v2 - W$ \\
        \midrule
        tfidf & $0.0$ & $0.6$ & $8.3$ & $-0.6$         \\
        openai & $0.0$ & $1.4$ & $9.8$ & $1.8$        \\
        \bottomrule
    \end{tabular}
    \caption{Ablation studies for Principled Samplers v2 showing the percent increase in aggregated wasted opportunity metric for each of the three ablated variants relative to corresponding baseline.}
    \label{tab:ablationv2}
\end{table}

\ignore{ 
\begin{figure}[t]
    \includegraphics[width=0.9\columnwidth]{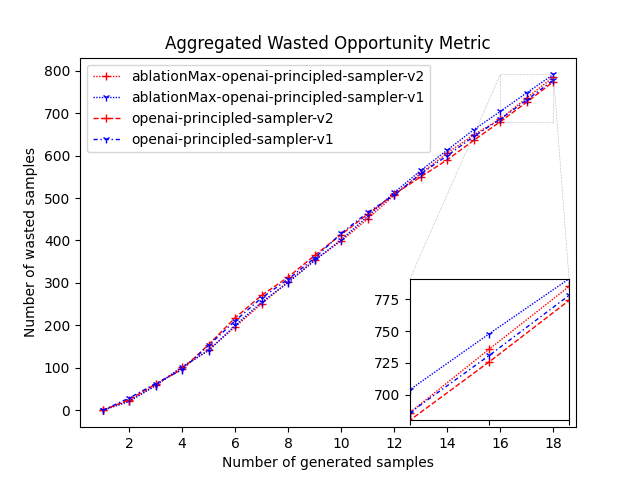}
    \includegraphics[width=0.9\columnwidth]{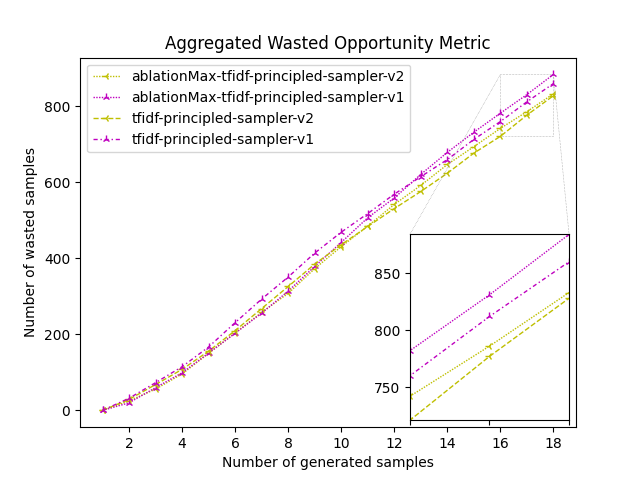}
    \caption{Removing the samples from the $Y$ set -- first with OpenAI embedding, second with TFIDF embedding.}
    \label{fig:ablationMax}
\end{figure}

Figure~\ref{fig:ablationMax} compares Principled Samplers v1 and v2 with their variants where $Y$ is replaced by random samples. These variants' names start with prefix \texttt{ablationMax}. We observe a small dip in performance of the ablated variants. 

Figure~\ref{fig:ablationMin} similarly compares Principled Samplers v1 and v2 with their variants where $Z$ is replaced by random samples. These variants' names start with prefix \texttt{ablationMin}. We observe a more significant drop in performance of the ablated variants. 

\begin{figure}[t]
    \includegraphics[width=0.9\columnwidth]{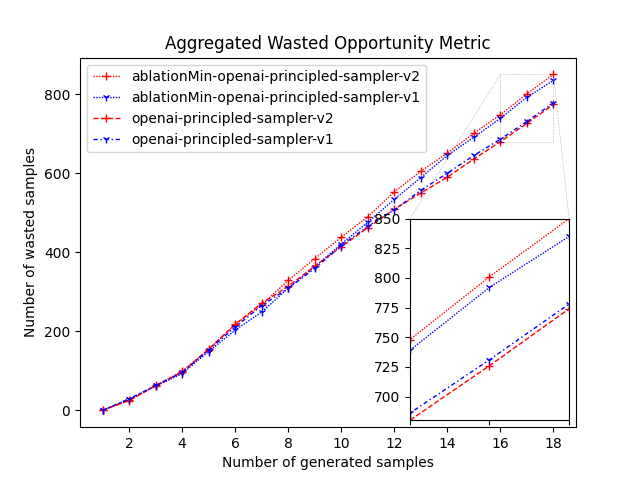}
    \includegraphics[width=0.9\columnwidth]{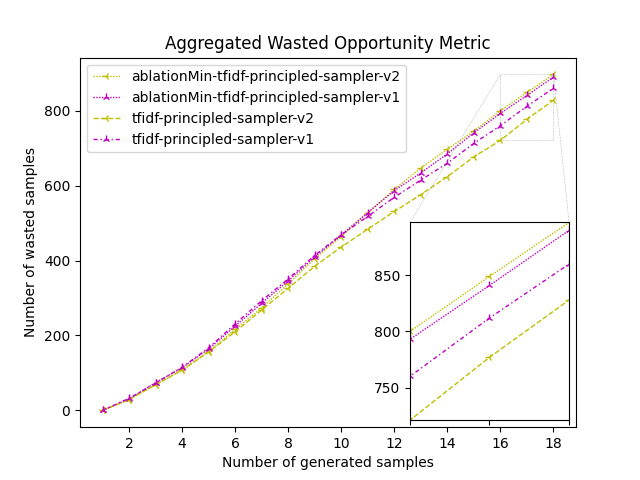}
    \caption{Removing the samples from the $Z$ set -- first with OpenAI embedding, second with TFIDF embedding.}
    \label{fig:ablationMin}
\end{figure}

Figure~\ref{fig:ablationZero} compares Principled Samplers v1 and v2 with their variants where $W$ is replaced by random samples. These variants' names start with prefix \texttt{ablationZero}.

\begin{figure}[t]
    \includegraphics[width=0.9\columnwidth]{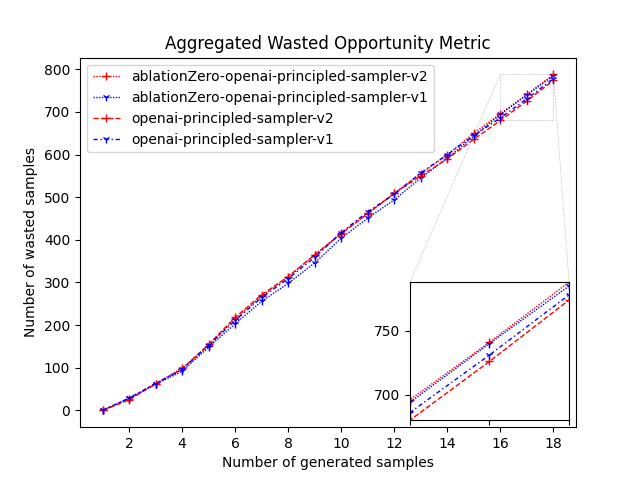}
    \includegraphics[width=0.9\columnwidth]{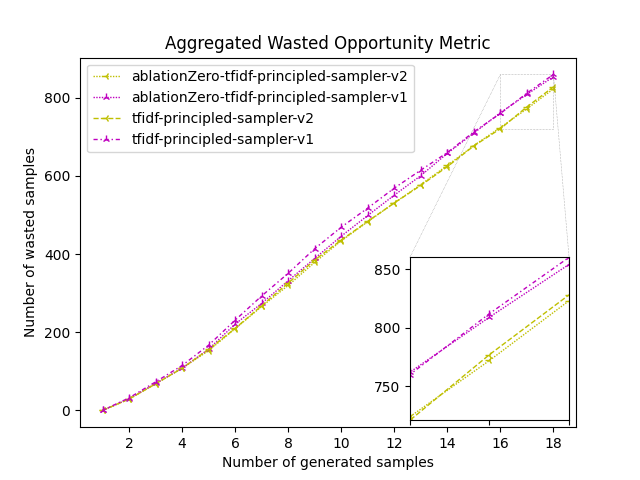}
    \caption{Removing the samples from the $W$ set -- first with OpenAI embedding, second with TFIDF embedding.}
    \label{fig:ablationZero}
\end{figure}
\endignore}

\subsection{Time Comparison}

\begin{figure}[t]
\includegraphics[trim=0 50 30 0,clip,width=0.9\columnwidth]{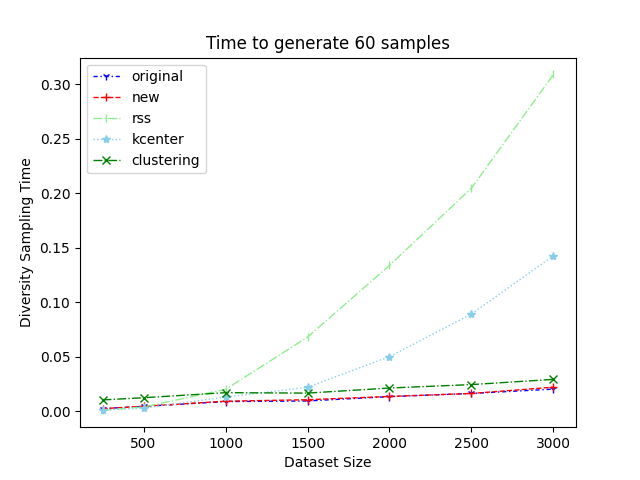} \includegraphics[trim=0 0 30 0,clip,width=0.9\columnwidth]{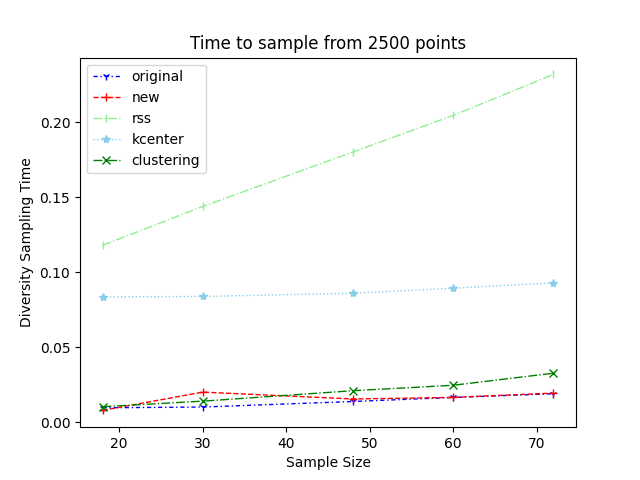}
    \caption{Time taken by different methods (a) to sample 60 from datasets of different sizes (b) to sample different number of items from a dataset with 2500 items.}
    \label{fig:timing}
\end{figure}

We finally compare the time taken by the different diverse sampling schemes to sample $60$ points as the size of the dataset grows from $250$ to $3000$. 
The top plot in Figure~\ref{fig:timing} shows that the proposed Principled Samplers v1 are the most efficient. This is surprising since PCA can be expensive; however, greedy k-center and reverse semantic search have a non-parallelizable iterative loop that makes them scale poorly. On the other hand, our proposed method can pick each sample almost independently of the other in parallel. We observe that clustering has time efficiency competitive with our methods.

The bottom chart in Figure~\ref{fig:timing} plots the time taken to sample $n$ points from a dataset of $2500$ points as $n$ grows from $18$ to $72$.  We again observe the same behavior as in the earlier plot. As a caveat, among all techniques, the implementation of our technique displayed the highest variance, which was possibly because the library PCA function picked different techniques based on the data. The plots were generated by doing multiple runs on all benchmarks and discarding outlier runs (for all techniques).

\section{Conclusion}

We presented a PCA-based approach for diversity sampling, which is task-agnostic and generates an ordered list of samples. We also defined the aggregated wasted opportunity metric for evaluating diversity in the generated samples, and showed that our technique outperforms techniques described in the literature. Even more, our technique is time efficient as compared to existing techniques.

\ignore{ 

\subsubsection{Submissions}
The following instructions apply to submissions:
\begin{itemize}
\item If your track requires submissions to be anonymous, they must be fully anonymized as discussed in the Modifications for Blind Review subsection below; in this case, Acknowledgements and Contribution Statement sections are not allowed.

\item If your track requires non-anonymous submissions, you should provide all author information at the time of submission, just as for camera-ready papers (see below); Acknowledgements and Contribution Statement sections are allowed, but optional.

\item Submissions must include line numbers to facilitate feedback in the review process . Enable line numbers by uncommenting the command {\tt \textbackslash{}linenumbers} in the preamble.

\item The limit on the number of  content pages is \emph{strict}. All papers exceeding the limits will be desk rejected.
\end{itemize}

\subsubsection{Camera-Ready Papers}
The following instructions apply to camera-ready papers:

\begin{itemize}
\item Authors and affiliations are mandatory. Explicit self-references are allowed. It is strictly forbidden to add authors not declared at submission time.

\item Acknowledgements and Contribution Statement sections are allowed, but optional.

\item Line numbering must be disabled. To achieve this, comment or disable {\tt \textbackslash{}linenumbers} in the preamble.

\item For some of the tracks, you can exceed the page limit by purchasing extra pages.
\end{itemize}

\subsection{Title and Author Information}

Center the title on the entire width of the page in a 14-point bold
font. The title must be capitalized using Title Case. For non-anonymous papers, author names and affiliations should appear below the title. Center author name(s) in 12-point bold font. On the following line(s) place the affiliations.

\subsubsection{Author Names}

Each author name must be followed by:
\begin{itemize}
    \item A newline {\tt \textbackslash{}\textbackslash{}} command for the last author.
    \item An {\tt \textbackslash{}And} command for the second to last author.
    \item An {\tt \textbackslash{}and} command for the other authors.
\end{itemize}

\subsubsection{Affiliations}

After all authors, start the affiliations section by using the {\tt \textbackslash{}affiliations} command.
Each affiliation must be terminated by a newline {\tt \textbackslash{}\textbackslash{}} command. Make sure that you include the newline after the last affiliation, too.

\subsubsection{Mapping Authors to Affiliations}

If some scenarios, the affiliation of each author is clear without any further indication (\emph{e.g.}, all authors share the same affiliation, all authors have a single and different affiliation). In these situations you don't need to do anything special.

In more complex scenarios you will have to clearly indicate the affiliation(s) for each author. This is done by using numeric math superscripts {\tt \$\{\^{}$i,j, \ldots$\}\$}. You must use numbers, not symbols, because those are reserved for footnotes in this section (should you need them). Check the authors definition in this example for reference.

\subsubsection{Emails}

This section is optional, and can be omitted entirely if you prefer. If you want to include e-mails, you should either include all authors' e-mails or just the contact author(s)' ones.

Start the e-mails section with the {\tt \textbackslash{}emails} command. After that, write all emails you want to include separated by a comma and a space, following the order used for the authors (\emph{i.e.}, the first e-mail should correspond to the first author, the second e-mail to the second author and so on).

You may ``contract" consecutive e-mails on the same domain as shown in this example (write the users' part within curly brackets, followed by the domain name). Only e-mails of the exact same domain may be contracted. For instance, you cannot contract ``person@example.com" and ``other@test.example.com" because the domains are different.

\subsubsection{Modifications for Blind Review}
When submitting to a track that requires anonymous submissions,
in order to make blind reviewing possible, authors must omit their
names, affiliations and e-mails. In place
of names, affiliations and e-mails, you can optionally provide the submission number and/or
a list of content areas. When referring to one's own work,
use the third person rather than the
first person. For example, say, ``Previously,
Gottlob~\shortcite{gottlob:nonmon} has shown that\ldots'', rather
than, ``In our previous work~\cite{gottlob:nonmon}, we have shown
that\ldots'' Try to avoid including any information in the body of the
paper or references that would identify the authors or their
institutions, such as acknowledgements. Such information can be added post-acceptance to be included in the camera-ready
version.
Please also make sure that your paper metadata does not reveal
the authors' identities.

\subsection{Abstract}

Place the abstract at the beginning of the first column 3$''$ from the
top of the page, unless that does not leave enough room for the title
and author information. Use a slightly smaller width than in the body
of the paper. Head the abstract with ``Abstract'' centered above the
body of the abstract in a 12-point bold font. The body of the abstract
should be in the same font as the body of the paper.

The abstract should be a concise, one-paragraph summary describing the
general thesis and conclusion of your paper. A reader should be able
to learn the purpose of the paper and the reason for its importance
from the abstract. The abstract should be no more than 200 words long.

\subsection{Text}

The main body of the text immediately follows the abstract. Use
10-point type in a clear, readable font with 1-point leading (10 on
11).

Indent when starting a new paragraph, except after major headings.

\subsection{Headings and Sections}

When necessary, headings should be used to separate major sections of
your paper. (These instructions use many headings to demonstrate their
appearance; your paper should have fewer headings.). All headings should be capitalized using Title Case.

\subsubsection{Section Headings}

Print section headings in 12-point bold type in the style shown in
these instructions. Leave a blank space of approximately 10 points
above and 4 points below section headings.  Number sections with
Arabic numerals.

\subsubsection{Subsection Headings}

Print subsection headings in 11-point bold type. Leave a blank space
of approximately 8 points above and 3 points below subsection
headings. Number subsections with the section number and the
subsection number (in Arabic numerals) separated by a
period.

\subsubsection{Subsubsection Headings}

Print subsubsection headings in 10-point bold type. Leave a blank
space of approximately 6 points above subsubsection headings. Do not
number subsubsections.

\paragraph{Titled paragraphs.} You should use titled paragraphs if and
only if the title covers exactly one paragraph. Such paragraphs should be
separated from the preceding content by at least 3pt, and no more than
6pt. The title should be in 10pt bold font and to end with a period.
After that, a 1em horizontal space should follow the title before
the paragraph's text.

In \LaTeX{} titled paragraphs should be typeset using
\begin{quote}
    {\tt \textbackslash{}paragraph\{Title.\} text} .
\end{quote}

\subsection{Special Sections}

\subsubsection{Appendices}
You may move some of the contents of the paper into one or more appendices that appear after the main content, but before references. These appendices count towards the page limit and are distinct from the supplementary material that can be submitted separately through CMT. Such appendices are useful if you would like to include highly technical material (such as a lengthy calculation) that will disrupt the flow of the paper. They can be included both in papers submitted for review and in camera-ready versions; in the latter case, they will be included in the proceedings (whereas the supplementary materials will not be included in the proceedings).
Appendices are optional. Appendices must appear after the main content.
Appendix sections must use letters instead of Arabic numerals. In \LaTeX,  you can use the {\tt \textbackslash{}appendix} command to achieve this followed by  {\tt \textbackslash section\{Appendix\}} for your appendix sections.

\subsubsection{Ethical Statement}

Ethical Statement is optional. You may include an Ethical Statement to discuss  the ethical aspects and implications of your research. The section should be titled \emph{Ethical Statement} and be typeset like any regular section but without being numbered. This section may be placed on the References pages.

Use
\begin{quote}
    {\tt \textbackslash{}section*\{Ethical Statement\}}
\end{quote}

\subsubsection{Acknowledgements}

Acknowledgements are optional. In the camera-ready version you may include an unnumbered acknowledgments section, including acknowledgments of help from colleagues, financial support, and permission to publish. This is not allowed in the anonymous submission. If present, acknowledgements must be in a dedicated, unnumbered section appearing after all regular sections but before references.  This section may be placed on the References pages.

Use
\begin{quote}
    {\tt \textbackslash{}section*\{Acknowledgements\}}
\end{quote}
to typeset the acknowledgements section in \LaTeX{}.

\subsubsection{Contribution Statement}

Contribution Statement is optional. In the camera-ready version you may include an unnumbered Contribution Statement section, explicitly describing the contribution of each of the co-authors to the paper. This is not allowed in the anonymous submission. If present, Contribution Statement must be in a dedicated, unnumbered section appearing after all regular sections but before references.  This section may be placed on the References pages.

Use
\begin{quote}
    {\tt \textbackslash{}section*\{Contribution Statement\}}
\end{quote}
to typeset the Contribution Statement section in \LaTeX{}.

\subsubsection{References}

The references section is headed ``References'', printed in the same
style as a section heading but without a number. A sample list of
references is given at the end of these instructions. Use a consistent
format for references. The reference list should not include publicly unavailable work.

\subsubsection{Order of Sections}
Sections should be arranged in the following order:
\begin{enumerate}
    \item Main content sections (numbered)
    \item Appendices (optional, numbered using capital letters)
    \item Ethical statement (optional, unnumbered)
    \item Acknowledgements (optional, unnumbered)
    \item Contribution statement (optional, unnumbered)
    \item References (required, unnumbered)
\end{enumerate}

\subsection{Citations}

Citations within the text should include the author's last name and
the year of publication, for example~\cite{gottlob:nonmon}.  Append
lowercase letters to the year in cases of ambiguity.  Treat multiple
authors as in the following examples:~\cite{abelson-et-al:scheme}
or~\cite{bgf:Lixto} (for more than two authors) and
\cite{brachman-schmolze:kl-one} (for two authors).  If the author
portion of a citation is obvious, omit it, e.g.,
Nebel~\shortcite{nebel:jair-2000}.  Collapse multiple citations as
follows:~\cite{gls:hypertrees,levesque:functional-foundations}.
\nocite{abelson-et-al:scheme}
\nocite{bgf:Lixto}
\nocite{brachman-schmolze:kl-one}
\nocite{gottlob:nonmon}
\nocite{gls:hypertrees}
\nocite{levesque:functional-foundations}
\nocite{levesque:belief}
\nocite{nebel:jair-2000}

\subsection{Footnotes}

Place footnotes at the bottom of the page in a 9-point font.  Refer to
them with superscript numbers.\footnote{This is how your footnotes
    should appear.} Separate them from the text by a short
line.\footnote{Note the line separating these footnotes from the
    text.} Avoid footnotes as much as possible; they interrupt the flow of
the text.

\section{Illustrations}

Place all illustrations (figures, drawings, tables, and photographs)
throughout the paper at the places where they are first discussed,
rather than at the end of the paper.

They should be floated to the top (preferred) or bottom of the page,
unless they are an integral part
of your narrative flow. When placed at the bottom or top of
a page, illustrations may run across both columns, but not when they
appear inline.

Illustrations must be rendered electronically or scanned and placed
directly in your document. They should be cropped outside \LaTeX{},
otherwise portions of the image could reappear during the post-processing of your paper.
When possible, generate your illustrations in a vector format.
When using bitmaps, please use 300dpi resolution at least.
All illustrations should be understandable when printed in black and
white, albeit you can use colors to enhance them. Line weights should
be 1/2-point or thicker. Avoid screens and superimposing type on
patterns, as these effects may not reproduce well.

Number illustrations sequentially. Use references of the following
form: Figure 1, Table 2, etc. Place illustration numbers and captions
under illustrations. Leave a margin of 1/4-inch around the area
covered by the illustration and caption.  Use 9-point type for
captions, labels, and other text in illustrations. Captions should always appear below the illustration.

\section{Tables}

Tables are treated as illustrations containing data. Therefore, they should also appear floated to the top (preferably) or bottom of the page, and with the captions below them.

\begin{table}
    \centering
    \begin{tabular}{lll}
        \hline
        Scenario  & $\delta$ & Runtime \\
        \hline
        Paris     & 0.1s     & 13.65ms \\
        Paris     & 0.2s     & 0.01ms  \\
        New York  & 0.1s     & 92.50ms \\
        Singapore & 0.1s     & 33.33ms \\
        Singapore & 0.2s     & 23.01ms \\
        \hline
    \end{tabular}
    \caption{Latex default table}
    \label{tab:plain}
\end{table}

\begin{table}
    \centering
    \begin{tabular}{lrr}
        \toprule
        Scenario  & $\delta$ (s) & Runtime (ms) \\
        \midrule
        Paris     & 0.1          & 13.65        \\
                  & 0.2          & 0.01         \\
        New York  & 0.1          & 92.50        \\
        Singapore & 0.1          & 33.33        \\
                  & 0.2          & 23.01        \\
        \bottomrule
    \end{tabular}
    \caption{Booktabs table}
    \label{tab:booktabs}
\end{table}

If you are using \LaTeX, you should use the {\tt booktabs} package, because it produces tables that are better than the standard ones. Compare Tables~\ref{tab:plain} and~\ref{tab:booktabs}. The latter is clearly more readable for three reasons:

\begin{enumerate}
    \item The styling is better thanks to using the {\tt booktabs} rulers instead of the default ones.
    \item Numeric columns are right-aligned, making it easier to compare the numbers. Make sure to also right-align the corresponding headers, and to use the same precision for all numbers.
    \item We avoid unnecessary repetition, both between lines (no need to repeat the scenario name in this case) as well as in the content (units can be shown in the column header).
\end{enumerate}

\section{Formulas}

IJCAI's two-column format makes it difficult to typeset long formulas. A usual temptation is to reduce the size of the formula by using the {\tt small} or {\tt tiny} sizes. This doesn't work correctly with the current \LaTeX{} versions, breaking the line spacing of the preceding paragraphs and title, as well as the equation number sizes. The following equation demonstrates the effects (notice that this entire paragraph looks badly formatted, and the line numbers no longer match the text):
\begin{tiny}
    \begin{equation}
        x = \prod_{i=1}^n \sum_{j=1}^n j_i + \prod_{i=1}^n \sum_{j=1}^n i_j + \prod_{i=1}^n \sum_{j=1}^n j_i + \prod_{i=1}^n \sum_{j=1}^n i_j + \prod_{i=1}^n \sum_{j=1}^n j_i
    \end{equation}
\end{tiny}%

Reducing formula sizes this way is strictly forbidden. We {\bf strongly} recommend authors to split formulas in multiple lines when they don't fit in a single line. This is the easiest approach to typeset those formulas and provides the most readable output%
\begin{align}
    x = & \prod_{i=1}^n \sum_{j=1}^n j_i + \prod_{i=1}^n \sum_{j=1}^n i_j + \prod_{i=1}^n \sum_{j=1}^n j_i + \prod_{i=1}^n \sum_{j=1}^n i_j + \nonumber \\
    +   & \prod_{i=1}^n \sum_{j=1}^n j_i.
\end{align}%

If a line is just slightly longer than the column width, you may use the {\tt resizebox} environment on that equation. The result looks better and doesn't interfere with the paragraph's line spacing: %
\begin{equation}
    \resizebox{.91\linewidth}{!}{$
            \displaystyle
            x = \prod_{i=1}^n \sum_{j=1}^n j_i + \prod_{i=1}^n \sum_{j=1}^n i_j + \prod_{i=1}^n \sum_{j=1}^n j_i + \prod_{i=1}^n \sum_{j=1}^n i_j + \prod_{i=1}^n \sum_{j=1}^n j_i
        $}.
\end{equation}%

This last solution may have to be adapted if you use different equation environments, but it can generally be made to work. Please notice that in any case:

\begin{itemize}
    \item Equation numbers must be in the same font and size as the main text (10pt).
    \item Your formula's main symbols should not be smaller than {\small small} text (9pt).
\end{itemize}

For instance, the formula
\begin{equation}
    \resizebox{.91\linewidth}{!}{$
            \displaystyle
            x = \prod_{i=1}^n \sum_{j=1}^n j_i + \prod_{i=1}^n \sum_{j=1}^n i_j + \prod_{i=1}^n \sum_{j=1}^n j_i + \prod_{i=1}^n \sum_{j=1}^n i_j + \prod_{i=1}^n \sum_{j=1}^n j_i + \prod_{i=1}^n \sum_{j=1}^n i_j
        $}
\end{equation}
would not be acceptable because the text is too small.

\section{Examples, Definitions, Theorems and Similar}

Examples, definitions, theorems, corollaries and similar must be written in their own paragraph. The paragraph must be separated by at least 2pt and no more than 5pt from the preceding and succeeding paragraphs. They must begin with the kind of item written in 10pt bold font followed by their number (e.g.: {\bf Theorem 1}),
optionally followed by a title/summary between parentheses in non-bold font and ended with a period (in bold).
After that the main body of the item follows, written in 10 pt italics font (see below for examples).

In \LaTeX{} we strongly recommend that you define environments for your examples, definitions, propositions, lemmas, corollaries and similar. This can be done in your \LaTeX{} preamble using \texttt{\textbackslash{newtheorem}} -- see the source of this document for examples. Numbering for these items must be global, not per-section (e.g.: Theorem 1 instead of Theorem 6.1).

\begin{example}[How to write an example]
    Examples should be written using the example environment defined in this template.
\end{example}

\begin{theorem}
    This is an example of an untitled theorem.
\end{theorem}

You may also include a title or description using these environments as shown in the following theorem.

\begin{theorem}[A titled theorem]
    This is an example of a titled theorem.
\end{theorem}

\section{Proofs}

Proofs must be written in their own paragraph(s) separated by at least 2pt and no more than 5pt from the preceding and succeeding paragraphs. Proof paragraphs should start with the keyword ``Proof." in 10pt italics font. After that the proof follows in regular 10pt font. At the end of the proof, an unfilled square symbol (qed) marks the end of the proof.

In \LaTeX{} proofs should be typeset using the \texttt{\textbackslash{proof}} environment.

\begin{proof}
    This paragraph is an example of how a proof looks like using the \texttt{\textbackslash{proof}} environment.
\end{proof}

\section{Algorithms and Listings}

Algorithms and listings are a special kind of figures. Like all illustrations, they should appear floated to the top (preferably) or bottom of the page. However, their caption should appear in the header, left-justified and enclosed between horizontal lines, as shown in Algorithm~\ref{alg:algorithm}. The algorithm body should be terminated with another horizontal line. It is up to the authors to decide whether to show line numbers or not, how to format comments, etc.

In \LaTeX{} algorithms may be typeset using the {\tt algorithm} and {\tt algorithmic} packages, but you can also use one of the many other packages for the task.

\begin{algorithm}[tb]
    \caption{Example algorithm}
    \label{alg:algorithm}
    \textbf{Input}: Your algorithm's input\\
    \textbf{Parameter}: Optional list of parameters\\
    \textbf{Output}: Your algorithm's output
    \begin{algorithmic}[1] 
        \STATE Let $t=0$.
        \WHILE{condition}
        \STATE Do some action.
        \IF {conditional}
        \STATE Perform task A.
        \ELSE
        \STATE Perform task B.
        \ENDIF
        \ENDWHILE
        \STATE \textbf{return} solution
    \end{algorithmic}
\end{algorithm}

\section{\LaTeX{} and Word Style Files}\label{stylefiles}

The \LaTeX{} and Word style files are available on the IJCAI--25
website, \url{https://2025.ijcai.org/}.
These style files implement the formatting instructions in this
document.

The \LaTeX{} files are {\tt ijcai25.sty} and {\tt ijcai25.tex}, and
the Bib\TeX{} files are {\tt named.bst} and {\tt ijcai25.bib}. The
\LaTeX{} style file is for version 2e of \LaTeX{}, and the Bib\TeX{}
style file is for version 0.99c of Bib\TeX{} ({\em not} version
0.98i). .

The Microsoft Word style file consists of a single file, {\tt
        ijcai25.docx}. 

These Microsoft Word and \LaTeX{} files contain the source of the
present document and may serve as a formatting sample.

Further information on using these styles for the preparation of
papers for IJCAI--25 can be obtained by contacting {\tt
        proceedings@ijcai.org}.

\appendix

\section*{Ethical Statement}

There are no ethical issues.

\section*{Acknowledgments}

The preparation of these instructions and the \LaTeX{} and Bib\TeX{}
files that implement them was supported by Schlumberger Palo Alto
Research, AT\&T Bell Laboratories, and Morgan Kaufmann Publishers.
Preparation of the Microsoft Word file was supported by IJCAI.  An
early version of this document was created by Shirley Jowell and Peter
F. Patel-Schneider.  It was subsequently modified by Jennifer
Ballentine, Thomas Dean, Bernhard Nebel, Daniel Pagenstecher,
Kurt Steinkraus, Toby Walsh, Carles Sierra, Marc Pujol-Gonzalez,
Francisco Cruz-Mencia and Edith Elkind.

\endignore}

\bibliographystyle{named}
\bibliography{ijcai25}

\end{document}